%% file: main.tex
\documentclass{article}

\usepackage[final]{neurips_2022}

\usepackage[utf8]{inputenc} 
\usepackage[T1]{fontenc}    
\usepackage{hyperref}       
\usepackage{url}            
\usepackage{booktabs}       
\usepackage{amsfonts}       
\usepackage{nicefrac}       
\usepackage{microtype}      
\usepackage{xcolor}         
\usepackage{multirow,multicol}
\usepackage{color,tabularx, colortbl,amsmath,bm,arydshln}
\usepackage{enumitem,kantlipsum,arydshln}
\usepackage{booktabs,url,tcolorbox,mathtools} \usepackage{newtxmath}
\usepackage{comment}

\newcounter{todocnt}

\author{
Shrestha Mohanty$^1$\thanks{Equal contribution} , Negar Arabzadeh$^{2*}$, Milagro Teruel$^1$, Yuxuan Sun$^3$,\\
\textbf{Artem Zholus$^4$, Alexey Skrynnik$^{3,4}$, Mikhail Burtsev$^{4,5}$, Kavya Srinet$^3$},\\ \textbf{Aleksandr Panov$^{4,5}$, Arthur Szlam$^3$, Marc-Alexandre Côté$^1$, Julia Kiseleva$^1$}\\
Microsoft Research$^1$, University of Waterloo$^2$, Meta Research$^3$,\\ Moscow Institute of Physics and Technology$^4$, Artificial Intelligence
Research Institute$^5$\\
}

\title{Collecting Interactive Multi-modal Datasets for Grounded Language Understanding}

\begin{document}

\maketitle
\begin{abstract}Human intelligence can remarkably adapt quickly to new tasks and environments. Starting from a very young age, humans acquire new skills and learn how to solve new tasks either by imitating the behavior of others or by following provided natural language instructions. To facilitate research which can enable similar capabilities in machines, we made the following contributions (1)~formalized the collaborative embodied agent using natural language task; (2)~developed a tool for extensive and scalable data collection; and (3)~collected the first dataset for interactive grounded language understanding.
\end{abstract}

\section{Introduction}
\input{01-introduction.tex}

\section{Collecting interactive data}
\input{02-data-collection.tex}

\section{Dataset and collection Tool description}
\input{03-stat.tex}

\section{Conclusions}
\input{04-conclusion.tex}

\bibliographystyle{plainnat}
\bibliography{ref}
\appendix
\input{appendix}

\end{document}

%% file: 01-introduction.tex
\label{sec:intro}

Humans learn and acquire new skills by imitating others or by following instructions in natural language~\citep{an_imitation_1988, council_how_1999}. Studies in developmental psychology have shown evidence of natural language communication being an effective method for imparting generic knowledge to individuals as young as infants~\citep{csibra2009natural}. This form of learning can even accelerate the acquisition of new skills by avoiding trial-and-error when learning occurs only from observations~\citep{thomaz2019interaction}. 
Inspired by this, the AI research community has attempted to develop grounded interactive \textit{}{embodied agents}~\cite{kiseleva2021neurips,kiseleva2022interactive, kojima2021continual} that are capable of engaging in natural back-and-forth dialog with humans to assist them in completing real-world tasks~\citep{winograd1971procedures,narayan2017towards, levinson2019tom,chen2020ask,abramson2020imitating}. Notably, the agent needs to understand when to initiate feedback requests if communication fails or instructions are not clear and requires learning new domain-specific vocabulary~\citep{aliannejadi2020convai3, aliannejadi2021building,rao2018learning, narayan2019collaborative, jayannavar-etal-2020-learning}.
Despite all these efforts, the task is far from solved. One of the biggest challenges is lack of general approaches for data collection and human-in-the-loop evaluation, where the collaborative agent is paired with a human.

In this paper, we present the following contributions:  
\begin{enumerate}[leftmargin=*,label=\textbf{C\arabic*},nosep]
  \item formalization of a task that allows studying collaborative embodied agents that use natural language for communication (Section~\ref{sec:task});  
  \item a large dataset to enable research for the above task;
  \item comprehensive open-sourced tooling for scalable data collection and to allow reproduction of experiments and results;
\end{enumerate}


%% file: 02-data-collection.tex
\label{sec:task}
The goal of the Collaborative Building Task, similar to~\cite{kiseleva2022interactive,jayannavar-etal-2020-learning, narayan2019collaborative}, is to train interactive embodied agents that learn to solve a task while provided with grounded natural language instructions in a collaborative environment.
By \emph{interactive agent}, we mean that the agent can: (1) follow the instructions provided in natural language in relation to the current world correctly, (2) ask for clarification in case of uncertainty, and (3) quickly adapt to newly acquired skills.

~\citet{narayan2019collaborative} has proposed the following setup: An Architect is provided with a target structure that needs to be built by the Builder. The Architect provides instructions to the Builder on how to create the target structure, and the Builder can ask clarifying questions to the Architect if an instruction is unclear~\cite{zhang-etal-2021-learning}. This dialog happens by means of a chat interface. The Architect is invisible to the Builder, while the Architect can see the actions of the Builder. ~\citet{narayan2019collaborative}  required installing Microsoft’s Project Malmo~\cite{johnson2016malmo} client, which provides an API for Minecraft agents to chat, build, and the ability to save and load game states. This is used to collect multi-turn interactions between the Architect and the Builder collaboratively working towards the common goal of building a given target structure. However, the data collection setup is limited to lab-based studies, which prevents massive online data collection.

\subsection{Single-Turn Data Collection}
\label{sec:single-turn}
\begin{figure}[t]
    \centering
    \includegraphics[width=0.99\textwidth]{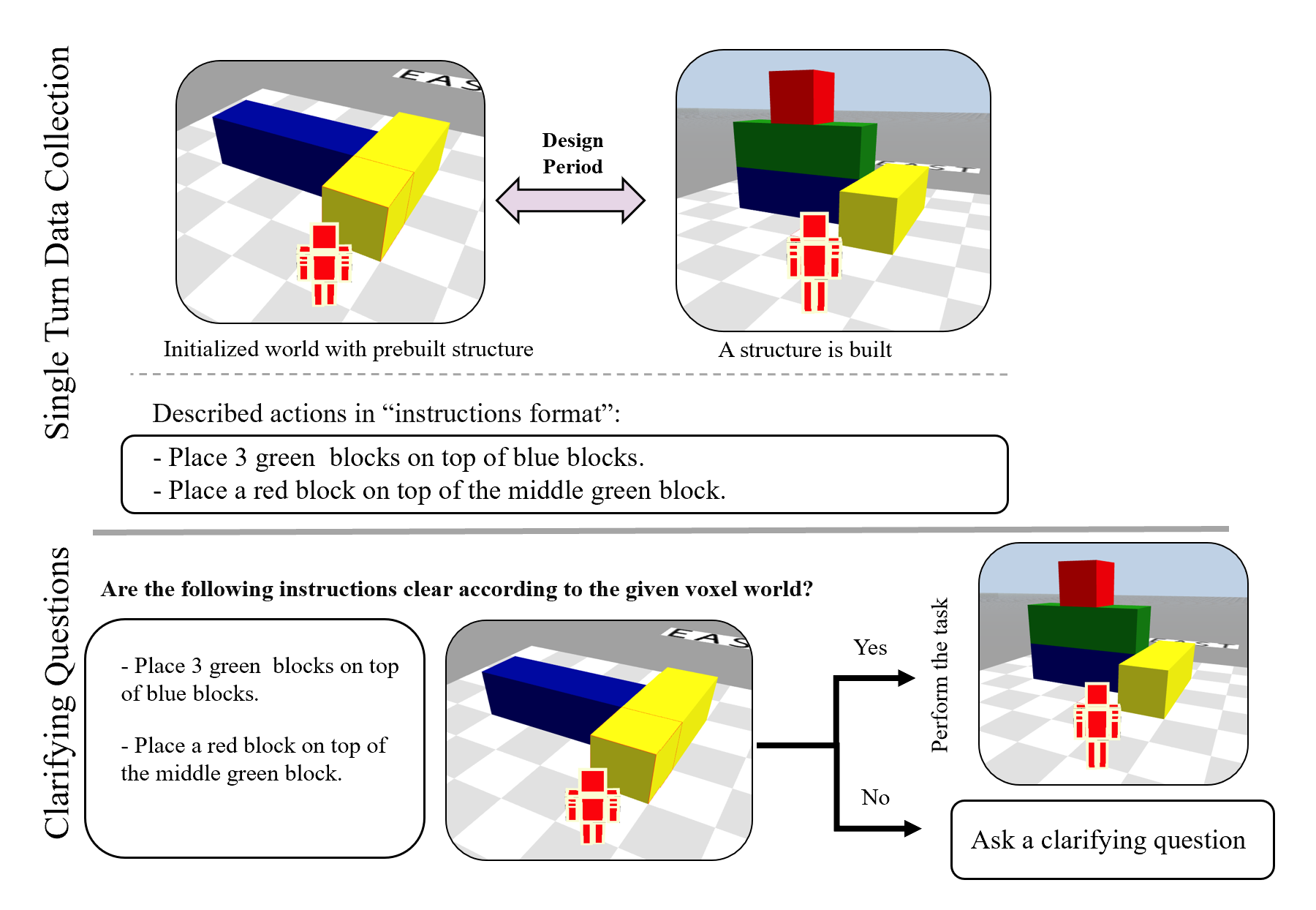}
    \caption{Overview of data collection for a single turn and clarifying questions}
    \label{fig:data-collection}
\end{figure}

\paragraph{Scaling up data collection} We have leveraged and extended the previously collected multi-turn interactions dataset in our work. For that, we modified the data collection strategy by removing the need for installing a local Minecraft client. We use an online crowdsourcing platform such as Amazon Mechanical Turk\footnote{\url{https://www.mturk.com/}} where we integrate our data collection tool, which leverages the CraftAssisft library~\cite{srinet-etal-2020-craftassist}. Our data collection tool can be easily plugged into the crowdsourcing platform enabling us to scale for participants quickly.

\paragraph{Simplifying the task} We further breakdown and simplify the multi-turn dialogues interactions to single-turn interactions. We do this by removing the added complexity of building a target structure and instead having an agent perform actions and provide instructions to another agent. We also leverage the multi-turn interactions data collected earlier to provide a starting point from which agents can build.
To elaborate, we design on the following setup, as shown in Figure~\ref{fig:data-collection}, for collecting data:
\begin{itemize}
    \item An interactive agent is dropped in the middle of an ongoing game where the structure is built partially. The partially completed game is retrieved from the multi-turn interactions dataset mentioned above.
    \item The agent is prompted to perform a sequence of actions for a duration of one minute.
    \item After which, the agent describes their performed set of actions in natural language, which will be displayed to another agent as an instruction.
    \item The next agent is shown the instruction and is asked to perform the steps mentioned in the instruction. If the instruction is not clear, the agent specifies it as thus and asks clarification questions.
\end{itemize}
This setup enables us to collect a dataset consisting of natural language instructions, actions performed based on those instructions, and a set of clarifying questions (described more in Section~\ref{sec:clarifying-questions}). Since the current data collection is limited to building-related tasks, the agent and builder are interchangeably used. Currently, we are working on enhancing the data collection tool to allow agents to perform various tasks, such as `grab,’ `bring,’ etc., to convert a builder into an actor. The dataset can be used to train models and agents towards solving tasks such as \emph{instruction generation} and \emph{building structures given natural language instructions}.

\subsection{Clarifying Questions}
\label{sec:clarifying-questions}
One of the advantages of collecting data in individual single turns is that the independence of every sample in the collected data will allow it to be used for different tasks more easily. In other words, it could be plugged into different settings, and each turn can be interpreted separately as a complete set of information for that turn. For instance, single-turn data provide a great host for collecting a comprehensive dataset for the \emph{Asking Clarifying Question} task. This has been studied in Information Retrieval, and Natural Language Processing domain to some extent ~\cite{aliannejadi2021building,aliannejadi2020convai3,AliannejadiSigir19,braslavski2017you,zamani2020generating}. While \textit{what to ask} has been relatively extensively studied~\cite{zamani2020generating,aliannejadi2021building,aliannejadi2019asking,braslavski2017you,rao2018learning,stoyanchev2014towards,de2003analysis,10.1007/978-3-030-99736-6_28,sekulic2021towards}, \textit{when to ask} clarifying questions in the case of unclear instructions or queries has been understudied and explored to limited extent. To the best of our knowledge ~\citet{aliannejadi2021building} were the first who explored both research questions i,e., \textit{when} and \textit{what} to ask in one work\footnote{https://github.com/aliannejadi/ClariQ}. The advantage of collecting such data in this way is that \emph {first}, it is close to real-world problems and applications, and \emph {second}, it has the potential to be used to evaluate the whole pipeline all at once. In other words, it can evaluate how often a system decided correctly to issue clarifying questions and further more if they are asking appropriate clarifying questions in that case.
While \emph{when to ask} clarifying questions are usually treated as a classification task, i.e., either the system asks a clarifying question or not, the second question can be answered in generation mode, i.e., a model generates a clarifying question, and the data can be used to  evaluate how close the generated questions are  to the target clarifying questions. As another option, the data can be used as a ranking problem in which the model is supposed to find the most relevant clarifying question from a corpus of all collected clarifying questions.

To collect such a dataset, we leveraged the instructions in Section~\ref{sec:single-turn}, and as shown in Figure~\ref{fig:data-collection}, we asked annotators if the instructions were clear. In case they found the instruction clear enough, we ask them to perform the instructions, i.e., try to continue building the structure given the clear instruction. Otherwise, when the instructions are decided as ambiguous, we ask the annotators to issue clarifying questions which might helps them to understand the instructions better. Figure~\ref{mturk} in the appendix illustrates the designed data collection web form. 

Next, we will provide a detailed overview of the collected dataset and tool.

%% file: 03-stat.tex
\label{sec:stat}
 \begin{table}[!t]
\centering
\caption{Statistics of single turn and clarifying questions collected data}
\begin{tabular}{llrrll}
 &  & \multicolumn{2}{c}{Instructions} & \multicolumn{2}{c}{Average Length (words)} \\
Section & \#Samples & \multicolumn{1}{l}{\#Clear} & \multicolumn{1}{l}{\#Ambigous} & \multicolumn{1}{c}{clarifying questions} & instructions \\ \hline
train & \multicolumn{1}{r}{6843} & 5951 & 892 & 12.12 & 18.51 \\
test & \multicolumn{1}{r}{1293} & 1129 & 164 & 11.99 & 18.08\\
\hline
\end{tabular}
\label{tab:table}
\end{table}

This section provides details of the collected single-turn and clarifying questions data.
As shown in Table~\ref{tab:table}, in total, after processing and cleaning the data, we have 8,136 single-turn data pairs of instructions and actions. Every single sample is randomly initialized with a pre-built structure from previously collected multi-turn interactions data. We filtered the data by some heuristic criteria, including but not limited to the given instruction must be in "English," and the length of the instructions should not very short. We manually evaluated the data during the curation stage and filtered out jobs that were annotated by the annotators who had low-quality instructions e.g., those who kept issuing the same instructions. As shown in Table~\ref{tab:table}, the filtered data, on average, has 18 words per instruction. This indicates that the instructions are descriptive enough for a one-minute building process. 

After collecting and filtering the single-turn data, we run a data annotation process for clarifying questions on the collected single-turn data as described in the lower section of Figure~\ref{fig:data-collection}. We kept the filtering heuristics used in single turn data collection for clarifying questions, i.e., the clarifying question must be in English and not too short; as well as included additional filtering criteria. For example, if annotators annotate the instructions as ambiguous, they must have issued a clarifying question. Otherwise, they would get a warning, and the task would be filtered out. This was to ensure that every instruction annotated as "not clear" is accompanied by at least one clarifying question. Similar to filtering single-turn data, we manually evaluated collected clarifying questions data frequently and blocked annotators that kept issuing similar clarifying questions. Out of 8,136 instructions , 1,056 (12.98\%) were annotated as \textit{Not Clear} and 7,080 (87.02\%) as \textit{Clear} instructions. The average length of clarifying questions is around 12 words, which indicates that the questions are specific.

Table~\ref{tab:unclear-questions} exemplifies some of the collected instructions which were annotated as "not clear" as well as their clarifying questions.  Consider the instruction issued by a crowdsourced annotator that says, \emph{"Place four blocks to the east of the highest block, horizontally"}. The corresponding clarifying question issued by another crowdsourced annotator was \emph{"Which color blocks?"}. This reflects that the data collection flow enables a natural, conversational way to get feedback about unclear instructions.

\paragraph{Open Source Collection Tool} We have released the data collection tool and the processed dataset on our GitHub repository\footnote{https://github.com/iglu-contest/iglu-dataset/}. The tool is run on the Amazon Mechanical Turk crowdsourcing platform to collect instructions, actions, gridworld states and clarifying questions in a scalable manner.\\ While the tool is currently limited to collecting single-turn instruction and action pairs, we plan to develop and release a multi-turn version that would enable multiple agents to interact and collaborate towards achieving a common goal through a shared task. 

\paragraph{Dataset Applications} The collected dataset can be used towards solving several Natural Language Processing and Reinforcement Learning tasks. Examples include:
\begin{itemize}
    \item Teaching an agent to build structures using natural language instructions.
    \item Generation of instruction given a completed structure.
    \item Building interactive agents that learn when and what feedback to seek when an unclear instruction is provided. As discussed in Section~\ref{sec:clarifying-questions}, this problem can be formulated into a classification task of when to ask and ranking problem of what to ask .
\end{itemize}

\paragraph{IGLU Challenge NeurIPS 2022} Above topics can help advance research in the areas of Natural Language Processing and Reinforcement Learning. Towards this, we set up the Interactive Grounded Language Understanding(IGLU)~\cite{IGLUProposal2022} competition which is part of NeurIPS 2022. The goal of this competition is to \emph {build embodied agents that learn to solve a task while provided with grounded natural language instructions in a collaborative environment}. The data we collected, is used to train models for the Reinforcement Learning and Natural Language Processing sub-tasks of the competition. Next, we elaborate on the baselines we set up for the Natural Language Processing task of the competition.

\paragraph{Baselines}
As a practical use case of the collected data for NLP community, we adapt the baselines from Clarifying Questions for Open-Domain Dialogue Systems (ClariQ) challenge on our collected data~\cite{aliannejadi2021building}. Given an initialized voxel world with pre-built structure and an instruction, the first goal is to determine whether any clarification question is needed.Thus, for \emph {when to ask Clarifying Questions}, we fine-tuned BERT~\cite{devlin2018bert} followed by a classification layer to predict if instructions are clear or not. This approach has shown to have promising performance on similar tasks~\cite{aliannejadi2021building,arabzadeh2022unsupervised}. Referring to Table~\ref{tab:table}, we see the majority of instructions are annotated as clear making the classification task of \emph{when to ask} clarifying questions even more challenging by having an unbalanced number of samples for each class.
After detecting ambiguous questions, \emph {what Clarifying Questions to ask} would be the main challenge. Inspired by ClariQ dataset ~\cite{aliannejadi2020convai3}, we curate a pool of clarification questions for each ambiguous instruction and the goal of this ranking-based task is to find the most relevant clarifying question for them. We adapted the well known BM25  to rank the clarifying questions in the question bank~\cite{robertson1995okapi}.

For the classification task i.e., When to ask clarifying question, the baseline got the F-1 score of  0.732.In addition, for the ranking task, since we have one clarification question per unclear instruction, we employed Mean Reciprocal Rank for evaluation purposes and BM25 achieved  0.341 in terms of MRR@20.   
In future, we would like to try question generation baselines on our dataset and report their performance on quality of generated clarifying questions.

\begin{table}[!t]
\centering
\caption{Examples of unclear instructions and their clarifying questions}
\scalebox{0.8}{
\begin{tabular}{|p{11cm}|p{5cm}|}
\hline
\\Instruction & Clarifying Question
\\[0.5em] \hline
Place four blocks to the east of the highest block, horizontally. & Which color blocks?\\[0.5em] \hline
Destroy 2 purple blocks and then build 3   green blocks diagonally. & Which two purple   blocks need to be destroyed? \\ \hline
Destroy the 3 stacked red blocks on the east side. Replace them with 3   stacked blue boxes & Which three of the four stacked red   blocks on the east side need to be destroyed? \\ \hline
Make a rectangle that is the width and height of the blue shape and fill   it in with purple blocks. & Which side I need to make the rectangle   is not clear \\ \hline
Facing South remove the rightmost purple block. Place a row of three   orange blocks to the left of the upper leftmost purple block. Place two   orange blocks above and below the leftmost orange block. & Which one of the rightmost blocks should   be removed? \\ \hline
facing north and purple green blocks will arrange on one by one. & Where would you like to place the purple   and green blocks exactly? \\ \hline
\end{tabular}}
\label{tab:unclear-questions}
\end{table}


%% file: 04-conclusion.tex
\label{sec:conclusion}
Studying interactive agents that have the ability of grounded language understanding is crucial to the development of the field. One of the current bottlenecks is a lack of an interactive platform for flexible data collection from multiple humans. In our work, we formalized and simplified an interactive, collaborative task and developed open-sourced tooling for flexible data collection that allows the easy plug-in to any crowdsourcing platform that makes it scalable. Moreover, we have collected an extensive dataset shared with the community to enable the study of grounded language understanding.

%% file: appendix.tex
\section{Appendix}
\label{sec:appendix}

\begin{figure}[!h]
    \centering
    \includegraphics[width=1\textwidth]{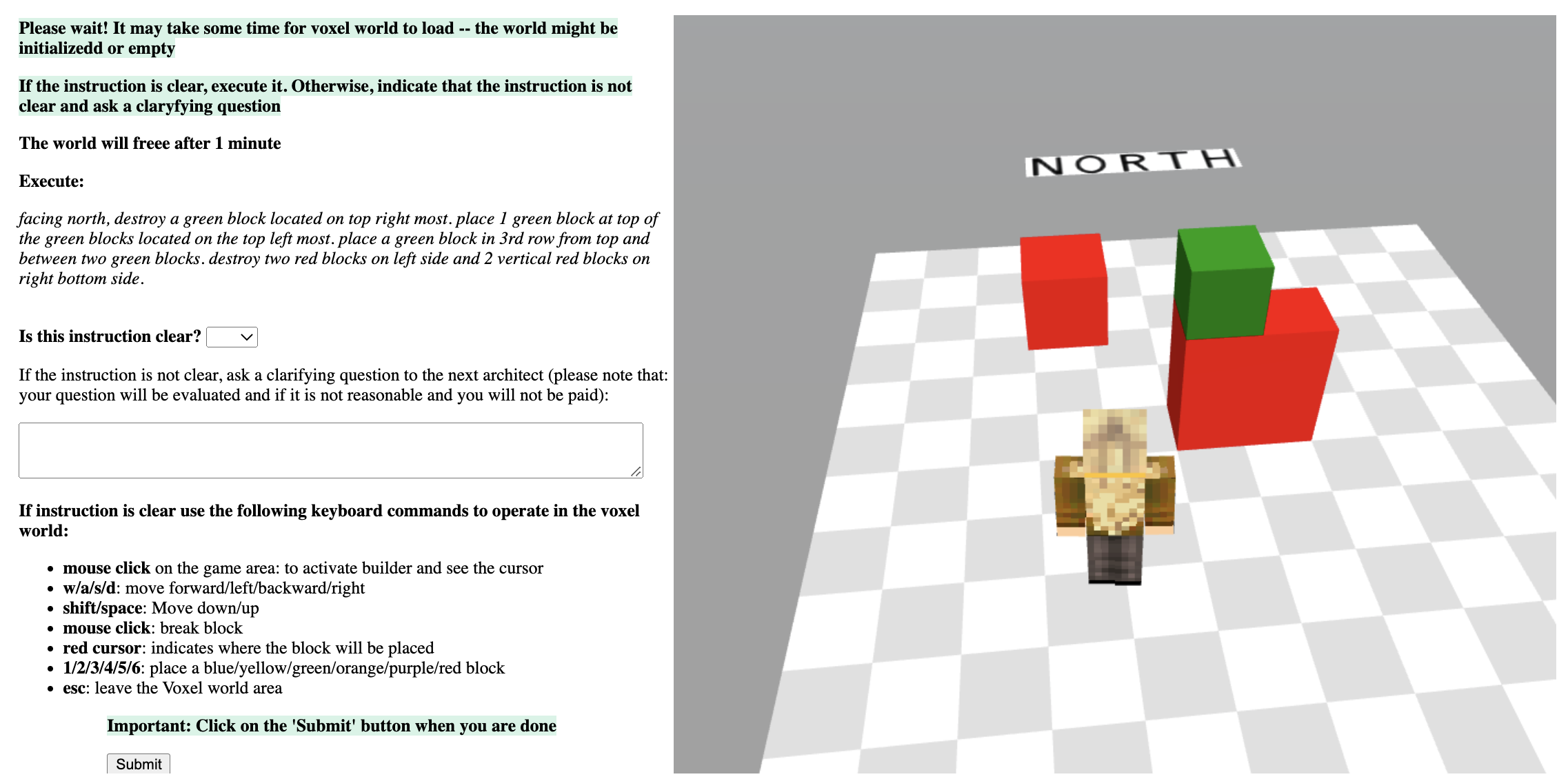}
    \caption{A web view of clarifying question data collection on Amazon Mechanical Turk }
    \label{mturk}
\end{figure}

%% file: main.bbl
\begin{thebibliography}{32}
\providecommand{\natexlab}[1]{#1}
\providecommand{\url}[1]{\texttt{#1}}
\expandafter\ifx\csname urlstyle\endcsname\relax
  \providecommand{\doi}[1]{doi: #1}\else
  \providecommand{\doi}{doi: \begingroup \urlstyle{rm}\Url}\fi

\bibitem[Abramson et~al.(2020)Abramson, Ahuja, Barr, Brussee, Carnevale,
  Cassin, Chhaparia, Clark, Damoc, Dudzik, et~al.]{abramson2020imitating}
Josh Abramson, Arun Ahuja, Iain Barr, Arthur Brussee, Federico Carnevale, Mary
  Cassin, Rachita Chhaparia, Stephen Clark, Bogdan Damoc, Andrew Dudzik, et~al.
\newblock Imitating interactive intelligence.
\newblock \emph{arXiv preprint arXiv:2012.05672}, 2020.

\bibitem[Aliannejadi et~al.(2019{\natexlab{a}})Aliannejadi, Zamani, Crestani,
  and Croft]{AliannejadiSigir19}
Mohammad Aliannejadi, Hamed Zamani, Fabio Crestani, and W.~Bruce Croft.
\newblock Asking clarifying questions in open-domain information-seeking
  conversations.
\newblock In \emph{International {ACM} {SIGIR} Conference on Research and
  Development in Information Retrieval (SIGIR)}, {SIGIR '19},
  2019{\natexlab{a}}.

\bibitem[Aliannejadi et~al.(2019{\natexlab{b}})Aliannejadi, Zamani, Crestani,
  and Croft]{aliannejadi2019asking}
Mohammad Aliannejadi, Hamed Zamani, Fabio Crestani, and W~Bruce Croft.
\newblock Asking clarifying questions in open-domain information-seeking
  conversations.
\newblock In \emph{Proceedings of the 42nd international acm sigir conference
  on research and development in information retrieval}, pages 475--484,
  2019{\natexlab{b}}.

\bibitem[Aliannejadi et~al.(2020)Aliannejadi, Kiseleva, Chuklin, Dalton, and
  Burtsev]{aliannejadi2020convai3}
Mohammad Aliannejadi, Julia Kiseleva, Aleksandr Chuklin, Jeff Dalton, and
  Mikhail Burtsev.
\newblock Convai3: Generating clarifying questions for open-domain dialogue
  systems (clariq).
\newblock \emph{arXiv preprint arXiv:2009.11352}, 2020.

\bibitem[Aliannejadi et~al.(2021)Aliannejadi, Kiseleva, Chuklin, Dalton, and
  Burtsev]{aliannejadi2021building}
Mohammad Aliannejadi, Julia Kiseleva, Aleksandr Chuklin, Jeff Dalton, and
  Mikhail Burtsev.
\newblock Building and evaluating open-domain dialogue corpora with clarifying
  questions.
\newblock In \emph{Proceedings of the 2021 Conference on Empirical Methods in
  Natural Language Processing}, pages 4473--4484, 2021.

\bibitem[An(1988)]{an_imitation_1988}
Meltzoff An.
\newblock Imitation, {Objects}, {Tools}, and the {Rudiments} of {Language} in
  {Human} {Ontogeny}, February 1988.
\newblock URL \url{https://pubmed.ncbi.nlm.nih.gov/23997403/}.
\newblock ISSN: 0393-9375 Issue: 1-2 Publisher: Hum Evol Volume: 3.

\bibitem[Arabzadeh et~al.(2022)Arabzadeh, Seifikar, and
  Clarke]{arabzadeh2022unsupervised}
Negar Arabzadeh, Mahsa Seifikar, and Charles~LA Clarke.
\newblock Unsupervised question clarity prediction through retrieved item
  coherency.
\newblock In \emph{Proceedings of the 31st ACM International Conference on
  Information \& Knowledge Management}, pages 3811--3816, 2022.

\bibitem[Braslavski et~al.(2017)Braslavski, Savenkov, Agichtein, and
  Dubatovka]{braslavski2017you}
Pavel Braslavski, Denis Savenkov, Eugene Agichtein, and Alina Dubatovka.
\newblock What do you mean exactly? analyzing clarification questions in cqa.
\newblock In \emph{Proceedings of the 2017 Conference on Conference Human
  Information Interaction and Retrieval}, pages 345--348, 2017.

\bibitem[Chen et~al.(2020)Chen, Gupta, and Marino]{chen2020ask}
Valerie Chen, Abhinav Gupta, and Kenneth Marino.
\newblock Ask your humans: Using human instructions to improve generalization
  in reinforcement learning.
\newblock \emph{arXiv preprint arXiv:2011.00517}, 2020.

\bibitem[Council(1999)]{council_how_1999}
National~Research Council.
\newblock \emph{How {People} {Learn}: {Brain}, {Mind}, {Experience}, and
  {School}: {Expanded} {Edition}}.
\newblock August 1999.
\newblock ISBN 978-0-309-07036-2.
\newblock \doi{10.17226/9853}.
\newblock URL
  \url{https://www.nap.edu/catalog/9853/how-people-learn-brain-mind-experience-and-school-expanded-edition}.

\bibitem[Csibra and Gergely(2009)]{csibra2009natural}
Gergely Csibra and Gy{\"o}rgy Gergely.
\newblock Natural pedagogy.
\newblock \emph{Trends in cognitive sciences}, 13\penalty0 (4):\penalty0
  148--153, 2009.

\bibitem[De~Boni and Manandhar(2003)]{de2003analysis}
Marco De~Boni and Suresh Manandhar.
\newblock An analysis of clarification dialogue for question answering.
\newblock In \emph{Proceedings of the 2003 human language technology conference
  of the north american chapter of the association for computational
  linguistics}, pages 48--55, 2003.

\bibitem[Devlin et~al.(2018)Devlin, Chang, Lee, and Toutanova]{devlin2018bert}
Jacob Devlin, Ming-Wei Chang, Kenton Lee, and Kristina Toutanova.
\newblock Bert: Pre-training of deep bidirectional transformers for language
  understanding.
\newblock In \emph{Conference of the North American Chapter of the Association
  for Computational Linguistics: Human Language Technologies (NAACL)}, 2018.

\bibitem[Jayannavar et~al.(2020)Jayannavar, Narayan-Chen, and
  Hockenmaier]{jayannavar-etal-2020-learning}
Prashant Jayannavar, Anjali Narayan-Chen, and Julia Hockenmaier.
\newblock Learning to execute instructions in a {M}inecraft dialogue.
\newblock In \emph{Proceedings of the 58th Annual Meeting of the Association
  for Computational Linguistics}, pages 2589--2602, Online, July 2020.
  Association for Computational Linguistics.
\newblock \doi{10.18653/v1/2020.acl-main.232}.
\newblock URL \url{https://www.aclweb.org/anthology/2020.acl-main.232}.

\bibitem[Johnson et~al.(2016)Johnson, Hofmann, Hutton, and
  Bignell]{johnson2016malmo}
Matthew Johnson, Katja Hofmann, Tim Hutton, and David Bignell.
\newblock The malmo platform for artificial intelligence experimentation.
\newblock In \emph{IJCAI}, pages 4246--4247. Citeseer, 2016.

\bibitem[Kiseleva et~al.(2021)Kiseleva, Li, Aliannejadi, Mohanty, ter Hoeve,
  Burtsev, Skrynnik, Zholus, Panov, Srinet, Szlam, Sun, Hofmann, Galley, and
  Awadallah]{kiseleva2021neurips}
Julia Kiseleva, Ziming Li, Mohammad Aliannejadi, Shrestha Mohanty, Maartje ter
  Hoeve, Mikhail Burtsev, Alexey Skrynnik, Artem Zholus, Aleksandr Panov, Kavya
  Srinet, Arthur Szlam, Yuxuan Sun, Katja Hofmann, Michel Galley, and Ahmed
  Awadallah.
\newblock Neurips 2021 competition iglu: Interactive grounded language
  understanding in a collaborative environment.
\newblock \emph{arXiv preprint arXiv:2110.06536}, 2021.

\bibitem[Kiseleva et~al.(2022{\natexlab{a}})Kiseleva, Li, Aliannejadi, Mohanty,
  ter Hoeve, Burtsev, Skrynnik, Zholus, Panov, Srinet,
  et~al.]{kiseleva2022interactive}
Julia Kiseleva, Ziming Li, Mohammad Aliannejadi, Shrestha Mohanty, Maartje ter
  Hoeve, Mikhail Burtsev, Alexey Skrynnik, Artem Zholus, Aleksandr Panov, Kavya
  Srinet, et~al.
\newblock Interactive grounded language understanding in a collaborative
  environment: Iglu 2021.
\newblock In \emph{NeurIPS 2021 Competitions and Demonstrations Track}, pages
  146--161. PMLR, 2022{\natexlab{a}}.

\bibitem[Kiseleva et~al.(2022{\natexlab{b}})Kiseleva, Skrynnik, Zholus,
  Mohanty, Arabzadeh, C{\^o}t{\'e}, Aliannejadi, Teruel, Li, Burtsev,
  et~al.]{IGLUProposal2022}
Julia Kiseleva, Alexey Skrynnik, Artem Zholus, Shrestha Mohanty, Negar
  Arabzadeh, Marc-Alexandre C{\^o}t{\'e}, Mohammad Aliannejadi, Milagro Teruel,
  Ziming Li, Mikhail Burtsev, et~al.
\newblock Iglu 2022: Interactive grounded language understanding in a
  collaborative environment at neurips 2022.
\newblock \emph{arXiv preprint arXiv:2205.13771}, 2022{\natexlab{b}}.

\bibitem[Kojima et~al.(2021)Kojima, Suhr, and Artzi]{kojima2021continual}
Noriyuki Kojima, Alane Suhr, and Yoav Artzi.
\newblock Continual learning for grounded instruction generation by observing
  human following behavior.
\newblock \emph{Transactions of the Association for Computational Linguistics},
  9:\penalty0 1303--1319, 2021.

\bibitem[Levinson(2019)]{levinson2019tom}
Stephen~C Levinson.
\newblock Tom m. mitchell, simon garrod, john e. laird, stephen c. levinson,
  and kenneth r. koedinger.
\newblock \emph{Interactive Task Learning: Humans, Robots, and Agents Acquiring
  New Tasks through Natural Interactions}, 26:\penalty0 9, 2019.

\bibitem[Narayan-Chen et~al.(2017)Narayan-Chen, Graber, Das, Islam, Dan,
  Natarajan, Doppa, Hockenmaier, Palmer, and Roth]{narayan2017towards}
Anjali Narayan-Chen, Colin Graber, Mayukh Das, Md~Rakibul Islam, Soham Dan,
  Sriraam Natarajan, Janardhan~Rao Doppa, Julia Hockenmaier, Martha Palmer, and
  Dan Roth.
\newblock Towards problem solving agents that communicate and learn.
\newblock In \emph{Proceedings of the First Workshop on Language Grounding for
  Robotics}, pages 95--103, 2017.

\bibitem[Narayan-Chen et~al.(2019)Narayan-Chen, Jayannavar, and
  Hockenmaier]{narayan2019collaborative}
Anjali Narayan-Chen, Prashant Jayannavar, and Julia Hockenmaier.
\newblock Collaborative dialogue in minecraft.
\newblock In \emph{Proceedings of the 57th Annual Meeting of the Association
  for Computational Linguistics}, pages 5405--5415, 2019.

\bibitem[Rao and Daum{\'e}~III(2018)]{rao2018learning}
Sudha Rao and Hal Daum{\'e}~III.
\newblock Learning to ask good questions: Ranking clarification questions using
  neural expected value of perfect information.
\newblock \emph{arXiv preprint arXiv:1805.04655}, 2018.

\bibitem[Robertson et~al.(1995)Robertson, Walker, Jones, Hancock-Beaulieu,
  Gatford, et~al.]{robertson1995okapi}
Stephen~E Robertson, Steve Walker, Susan Jones, Micheline~M Hancock-Beaulieu,
  Mike Gatford, et~al.
\newblock Okapi at trec-3.
\newblock \emph{Nist Special Publication Sp}, 109:\penalty0 109, 1995.

\bibitem[Sekuli{\'c} et~al.(2021)Sekuli{\'c}, Aliannejadi, and
  Crestani]{sekulic2021towards}
Ivan Sekuli{\'c}, Mohammad Aliannejadi, and Fabio Crestani.
\newblock Towards facet-driven generation of clarifying questions for
  conversational search.
\newblock In \emph{Proceedings of the 2021 ACM SIGIR International Conference
  on Theory of Information Retrieval}, pages 167--175, 2021.

\bibitem[Sekuli{\'{c}} et~al.(2022)Sekuli{\'{c}}, Aliannejadi, and
  Crestani]{10.1007/978-3-030-99736-6_28}
Ivan Sekuli{\'{c}}, Mohammad Aliannejadi, and Fabio Crestani.
\newblock Exploiting document-based features for clarification in
  conversational search.
\newblock In Matthias Hagen, Suzan Verberne, Craig Macdonald, Christin Seifert,
  Krisztian Balog, Kjetil N{\o}rv{\aa}g, and Vinay Setty, editors,
  \emph{Advances in Information Retrieval}, pages 413--427, Cham, 2022.
  Springer International Publishing.
\newblock ISBN 978-3-030-99736-6.

\bibitem[Srinet et~al.(2020)Srinet, Jernite, Gray, and
  Szlam]{srinet-etal-2020-craftassist}
Kavya Srinet, Yacine Jernite, Jonathan Gray, and Arthur Szlam.
\newblock {C}raft{A}ssist instruction parsing: Semantic parsing for a
  voxel-world assistant.
\newblock In \emph{Proceedings of the 58th Annual Meeting of the Association
  for Computational Linguistics}, pages 4693--4714, Online, July 2020.
  Association for Computational Linguistics.
\newblock \doi{10.18653/v1/2020.acl-main.427}.
\newblock URL \url{https://www.aclweb.org/anthology/2020.acl-main.427}.

\bibitem[Stoyanchev et~al.(2014)Stoyanchev, Liu, and
  Hirschberg]{stoyanchev2014towards}
Svetlana Stoyanchev, Alex Liu, and Julia Hirschberg.
\newblock Towards natural clarification questions in dialogue systems.
\newblock In \emph{AISB symposium on questions, discourse and dialogue},
  volume~20, 2014.

\bibitem[Thomaz et~al.(2019)Thomaz, Lieven, Cakmak, Chai, Garrod, Gray,
  Levinson, Paiva, and Russwinkel]{thomaz2019interaction}
Andrea~L Thomaz, Elena Lieven, Maya Cakmak, Joyce~Y Chai, Simon Garrod, Wayne~D
  Gray, Stephen~C Levinson, Ana Paiva, and Nele Russwinkel.
\newblock Interaction for task instruction and learning.
\newblock In \emph{Interactive task learning: Humans, robots, and agents
  acquiring new tasks through natural interactions}, pages 91--110. MIT Press,
  2019.

\bibitem[Winograd(1971)]{winograd1971procedures}
Terry Winograd.
\newblock Procedures as a representation for data in a computer program for
  understanding natural language.
\newblock Technical report, MASSACHUSETTS INST OF TECH CAMBRIDGE PROJECT MAC,
  1971.

\bibitem[Zamani et~al.(2020)Zamani, Dumais, Craswell, Bennett, and
  Lueck]{zamani2020generating}
Hamed Zamani, Susan Dumais, Nick Craswell, Paul Bennett, and Gord Lueck.
\newblock Generating clarifying questions for information retrieval.
\newblock In \emph{Proceedings of The Web Conference 2020}, pages 418--428,
  2020.

\bibitem[Zhang et~al.(2021)Zhang, Jauhar, Kiseleva, White, and
  Roth]{zhang-etal-2021-learning}
Yi~Zhang, Sujay~Kumar Jauhar, Julia Kiseleva, Ryen White, and Dan Roth.
\newblock Learning to decompose and organize complex tasks.
\newblock In \emph{Proceedings of the 2021 Conference of the North American
  Chapter of the Association for Computational Linguistics: Human Language
  Technologies}, pages 2726--2735, 2021.

\end{thebibliography}
